Joaquín Gayoso-Cabada. Facultad de Informática. Universidad Complutense de Madrid. Madrid 28040.España. jgayoso@ucm.es
Antonio Sarasa-Cabezuelo. Facultad de Informática. Universidad Complutense de Madrid. Madrid 28040.España. asarasa@ucm.es
José Luis Sierra-Rodriguez. Facultad de Informática. Universidad Complutense de Madrid. Madrid 28040.España. jlsierra@ucm.es

# A review of annotation classification tools in the educational domain

**Abstract:** An annotation consists of a portion of information that is associated with a piece of content in order to explain something about the content or to add more information. The use of annotations as a tool in the educational field has positive effects on the learning process. The usual way to use this instrument is to provide students with contents, usually textual, with which they must associate annotations. In most cases this task is performed in groups of students who work collaboratively. This process encourages analysis and understanding of the contents since they have to understand them in order to annotate them, and also encourages teamwork. To facilitate its use, computer applications have been developed in recent decades that implement the annotation process and offer a set of additional functionalities. One of these functionalities is the classification of the annotations made. This functionality can be exploited in various ways in the learning process, such as guiding the students in the annotation process, providing information to the student about how the annotation process is done and to the teacher about how the students write and how they understand the content, as well as implementing other innovative educational processes. In this sense, the classification of annotations plays a critical role in the application of the annotation in the educational field. There are many studies of annotations, but most of them consider the classification aspect marginally only. This paper presents an initial study of the classification mechanisms used in the annotation tools, identifying four types of cases: absence of classification mechanisms, classification based on pre-established vocabularies, classification based on extensible vocabularies, and classification based on structured vocabularies.

## 1 Introduction

Annotations make it possible to enrich content with additional information that facilitates understanding. Traditionally, this process has been carried out by hand (Kawase et al, 2009). However, the introduction of computer science in this field has transformed the annotation process in several ways. First of all, the type of content that is annotated is digital (Chowdhury et al, 2002), (Schroeter et al, 2006). Second, it facilitates collaborative processes (Chen et al, 2014), (Hwang et al, 2007), (Johnson et al, 2010), (Mendenhall et al, 2010), (Nokelainen et al, 2003), (Nokelainen et al, 2005), (Su et al, 2010), so that a team of people can work at the same time on the annotation of the same contents. This favours higher annotation quality, as annotations are the result of different points of view (Jan et al, 2016). And thirdly, this type of tools offer services that facilitate the annotator's work, such as the possibility of using different types of annotations, classifying the annotations made, analysing the annotation process performed on a content, and analysing annotation styles.

Annotation has been used in different educational areas as an auxiliary tool in the learning process. The most common way to use annotations is to set up group annotation tasks under the supervision of a teacher who gives students instructions on how to perform the annotation process (Buendia et al, 2016). From the point of view of learning, annotations have positive effects as they strengthen and promote certain skills such as teamwork, reflective ability, and communication skills.

From the educational point of view, analysing the annotations made by students provides information about content comprehension (Cigarran et al, 2014), annotation styles, and intellectual maturity. In this sense, many annotation tools provide services oriented to exploit the annotations. The results of this exploitation are usually models that show how annotation types are related to each other, as well as other types of relationships that may exist among the annotations themselves (Novak et al, 2012). This information can be useful for creating annotation recommenders or for finding patterns in annotations. In this sense, a requirement to be able to exploit the annotations is the possibility of classifying the annotations.

This paper provides a review of the mechanisms that are used in annotation tools to classify the annotations made. The study has been limited to document annotations, since it is the most usual type of content. To carry out the study, an exhaustive bibliographic search in several current reviews of this topic was conducted, as well as searches in repositories of academic articles (ScienceDirect, ACM Digital Library, IEEE explore, and Google Scholar). As a result of the search, 38 different tools have been considered. Based on them, a classification of 4 types of tools has been defined, taking shared characteristics as a grouping criterion: 1) Tools with no classification mechanisms, 2) Tools that use pre-established vocabularies to classify, 3) Tools that use extensible vocabularies to classify, and 4) Tools that use structured vocabularies.

The article is structured as follows. Sections 2, 3, 4, and 5 present each type of tool. Their general features are shown, and examples are given. In section 6 new trends are described. In section 7, its use in the field of education is discussed. Finally, section 7 presents some conclusions and lines for future work. A preliminary version of this study can be found in (Gayoso-Cabada et al, 2018).

## 2 Tools without classification mechanisms

There is a considerable number of tools that lack explicit mechanisms to classify the annotations made or for later exploitation of the information obtained from the annotation process. In this sense, they are designed exclusively to make annotations and offer services to perform this process such as collaborative annotation or the possibility of using digital ink.

Some examples of this category are the following. Regarding the use of digital ink, we have WriteOn (Tron et al, 2006) or PaperCP (Liao et al, 2007), which supports annotations in digital ink of presentations in the classroom using tablets. Some examples of the ability to annotate collaboratively, and sharing content and annotations are Digital Reading Desk (Pearson et al, 2012), which enables collaborative annotation of ebooks based on a virtual desktop, Livenotes (Kam et al, 2005) which enables collaborative annotation of presentations in PowerPoint, and u-Annotate (Chatti et al, 2006), which supports annotation of web pages by hand. Table 1 shows the main characteristics of these tools.

**Table 1:** Tools without classification mechanisms

| Tool | Mechanisms | Content |
|---|---|---|
| WriteOn | Digital ink | Classroom presentations using tablet PCs |
| PaperCP | Digital ink | Classroom presentations using tablet PCs |
| Digital Reading Desk | Collaborative annotation | E-books |
| Livenotes | Collaborative annotation | PowerPoint presentations |
| u-Annotate | Handwritten annotation | Web pages |

## 3 Tools that use pre-established vocabularies.

These tools are characterised by having a set of tags that constitute a closed vocabulary with no structure, used to label the annotations. In general extending the vocabularies, or using tags different from the pre-established ones is not possible. There are two types of tools according to the aspect to be tagged. Firstly, there are the tools that tag the way in which content is annotated, putting the focus of interest on the style of annotation. Thus, these tools highlight aspects such as the using of underlining, bold type, emphasis and other elements related to the presentation of contents. Secondly, there are other types of tools that use tag vocabularies in which more attention is paid to the semantics of the annotated content, so that it is possible to determine whether the content is a discussion, an argument, etc. Section 3.1 describes the tools that use style tags and section 3.2 describes the tools that use semantic tags.

## 3.1  Tools that use style tags.

These tools are characterized by their offering a set of predefined tags to make annotations based on the ways of anno-tating the documents and the presentation attributes used, such as underlining, highlighting, explicit text, etc. Thus, they do not take into account the semantics of the annotation, but the presentational attributes of the contents.

Some examples of this approach are Adobe Reader and PDF Annotator (Shindo et al, 2018), which allow for different types of notes such as highlighted, inserted, and  deleted text, as well as sticky notes on pdf documents; Diigo (Deng et al, 2018), which allows users to highlight text fragments and add sticky notes (Lu et al, 2013);  CASE (Glover et al, 2004) which  allows users to add images, links, pop-ups and text notes; and CON2ANNO (Lin et al, 2014), which users them to highlight / underline text and add text notes, as well as add summaries based on existing notes to form integrated essays. Other tools benefit from information about the annotation activity to allow for more sophisticated uses, like VPen (Hwang et al, 2007), which allows annotators to highlight and underline text, as well as add text and voice; and IIAF (Assai, 2014) , which supports different annotation modes (underline, enclose or comment). Finally, there are tools that use clustering algorithms to group students into clusters in terms of their annotation activity, like as the tool described in (Chang et al, 2015This tool also changes annotation styles by underlining and highlighting, or by changing font sizes and styles. Table 2 summarises the main characteristics of these tools.

**Table 2:** Tools that uses style tags

| Tool | Style tags | Content | Other features |
|------|-----------|---------|----------------|
| Adobe Reader | Different types of notes such as highlighted, inserted or erased text, sticky notes | PDF documents | --- |
| Diigo | Highlight text fragments and add sticky notes | Web page | --- |
| CASE tool | Adding images, links, pop-ups and text notes | Web page | Collaborative annotation |
| CON2ANNO | Highlight / underline text and add text notes, as well as to add summaries based on existing notes to form integrated essays. | Web page | --- |
| VPen | Highlight and underline text, and add text and voice | Web page | Recommendation mechanism based on ranking users by the quantity of the notes provided |
| IIAF | fferent annotation modes (e.g., underlining, enclosure or commenting) | Web page | Automatic detection of the user's intention in order to automatically identify the intended annotation mode. |

## 3.2  Tools that use semantic tags.

These tools are characterised by their offering a set of predefined tags to classify the annotations based on their meaning / semantics. In this sense each tag represents a different semantic category.

Some examples of this approach are Highlight (Pereira et al, 2012), which uses two different semantic types (important and confusing) to make annotations, PAMS 2.0 (Su et al, 2010), which uses four predefined semantic types (definitions, comments, questions, and associations), MyNote (Chen et al, 2012), which includes four semantic categories (normal, question, answer, and discussion), Tafannote (Jan et al, 2016) that includes nine predefined semantic categories (com-ment, reference, positive and negative judgment, correction, question, example, confirm and refute), WCRAS-TQAFM (Jan et al, 2016) that uses five types of predefined semantic annotations (importance, quizzing, query, example, and summary) or the MADCOW tool (Bottoni et al, 2004) that offers a set of nine predefined types of semantic annotations (explanation, comment, question, integration, example, summary, solution, announcement, and memorandum). Finally, CRAS-RAID (Chen et al, 2014) provides two separate sets of semantic labels, one for labelling annotations (reasoning, discrimination, linking, summary, quizzing, explanation, and other) and another for labelling entries in discussion forums (reasoning, discrimination, quizzing, clarifications, debugging, and other). Table 3 shows the main characteristics of these tools.

**Table 2:** Tools that use semantic tags

| Tool | Semantic tags | Other features |
|---|---|---|
| Highlight | important and confusing | --- |
| MyNote | normal, question, answer, and discussion | It can be integrated in an LMS for annotating learning objects as well as external documents |
| PAMS 2.0 | definitions, comments, questions, associations | --- |
| Tafannote | comment, reference, positive and negative judgment, correction, question, example, confirm and refute | Collaborative annotation |
| WCRAS-TQAFM | importance, quizzing, query, example, and summary | Support for the automatic assessment of the quality of the annotations |
| CRAS-RAID | tagging notes (reasoning, discrimination, linking, summary, quizzing, explanation, and other) and tagging discussion entries (reasoning, discrimination, quizzing, clarification, debugging, and other) | Collaborative annotation |
| MADCOW | explanation, comment, question, integration, example, summary, solution, announcement and memorandum | Annotation of multimedia documents |

# 4 Tools that use extensible vocabularies (folksonomies)

Although these tools also have a set of tags that constitute a vocabulary without a structure, which is used to label the annotations, the main difference with respect to the previous ones is that they are not closed vocabularies, but rather they are built and extended using new tags created by the annotators themselves. That is why these tools have in common the fact that annotations are made collaboratively. These vocabularies are usually called folksonomies (Ovsiannikov et al, 1999). One of the advantages with respect to the previous vocabularies is that they are usually better suited to the annotation needs when they are created by the annotation experts and in a specific way for each annotation process.

Some examples of these tools are HyLighter (Lebow et al, 2005), which allows for the use if labels defined by the user; Annotation Studio (Paradis et al, 2016), which makes it possible to use sets of labels introduced by teachers in order to be used by students; A.nnotate (Anagnostopoulou et al, 2009), (Tseng et al, 2015), which makes it possible allows to classify free-text notes using labels or to create annotations whose content consists only of labels created by the users themselves; Note-taking (Kim et al, 2014) , which uses clouds of labels created by the students used to classify the annotations made on electronic books,; OATS (Bateman et al, 2007), (Bateman et al, 2006), which facilitates searches in collections of documents in terms of the associated annotations using non-prefixed vocabularies;SpreadCrumbs (Kawase et al, 2009), (Kawase et al, 2010), which enables the annotation of documents with discussion trails and their classification into user-defined topics; and Tsaap-Notes (Silvestre et al, 2014), which uses a classification scheme based on user-defined hashtags. Table 4 shows the main characteristics of these tools.

**Table 4:** Tools that use folksonomies

| Tool | Annotation features | Other features |
|---|---|---|
| HyLighter | User-defined tags | --- |
| Annotation Studio | User-defined tags and repertoires of tags introduced by instructors | --- |
| A.nnotate | User-defined tags and annotations whose content consists only of tags | --- |
| Note-taking | Tag clouds to classify their notes on e-books | Specifically oriented to the annotation of e-books |
| OATS tool | User-defined tags in order to browse in collections of digital documents | --- |

| Tool | Annotation features | Other features |
|---|---|---|
| SpreadCrumbs | Classification using user-defined topics | Promotes the annotation of documents with discussion trails. |
| Tsaap-Notes | Classification schema based on user-defined *hashtags*. | Adopts a micro-blogging paradigm for supporting annotation |

# 5  Tools that use structured vocabularies

These tools provide sets of predefined tags to make annotations. Their main characteristic is that the tags used are related to each other. In this sense it is possible to find taxonomies, thesauri and ontologies (Kalboussi A et al, 2015). Taxonomies present a hierarchical scheme of tags organized into categories and subcategories so that the structure of the vocabulary is tree-like. Thesauri are sets of tags in which synonymy, composition and association relationships have been defined in such a way that the structure of the vocabulary takes the form of a graph. Finally, ontologies are collections of tags in which relationships have been defined that represent properties that can be expressed through triplets of the subject-predicate-object type. This paper focuses on the tools that use ontologies, because of the three approaches it is the most flexible to perform annotation, as the tools makes it possible to select the ontology that best adapts to the semantic particularities of the annotated content, It is also the type of tool with the greatest semantic richness, as, by associating concepts to a content, all the relationships maintained are also inherited.

Some tools that follow this paradigm are loomp (Hinze et al, 2012), (Luczak-Rösch et al, 2009), which makes it possible to select relevant text fragments and annotate them with concepts taken from various ontologies so that the annotations become instances of the selected concepts; WebAnnot (Azouaou et al, 2013), which uses an ontology of domain annotation objectives and an ontology of document annotation objectives; DLNotes (Rocha et al, 2009), which allows for the annotation of text fragments with concept examples instead of the concepts themselves; MemoNote (Azozau et al, 2006),which allows users to incorporate three different types of ontologies (the ontology of pedagogical annotation objectives to make a specific pedagogical feature of the content explicit, the ontology of domain annotation objectives to note a certain aspect of the learning domain, and the ontology of annotation objectives of the document to record aspects related to the document itself); AeroDAML (Kogut et al, 2001)  maps proper nouns and common relationships onto classes and properties in DAML ontologies; AKTiveMedia (Chakravarthy et al,2006) (Bikakis et al, 2010) allows users to annotate  with ontology-based and free-text annotations; KIM (Popov et al, 2004), (Maliv et al, 2010), (Popov et al, 2003) uses an ontology, a knowledge base, an automatic Semantic Annotation, indexing, and a retrieval server; and @note (Handschuh  et al, 2002), (Corlosquet  et al, 2009), (Shreves  et al, 2011) makes it possible to coherently combine the semantic annotations of free text and ontology. Table 5 shows the main characteristics of these tools.

**Table 3:** Tools that use ontologies

| Tool | Ontologies | Other features |
|---|---|---|
| loomp | It annotates using concepts taken from several ontologies. | It allows grouping fragments in sets called mash-ups and semantically annotating these mash-ups with additional concepts and links |
| WebAnnot | Domain annotation objectives ontology and a document annotation objectives ontology | Excludes the *pedagogy annotation objectives* one |
| DLNotes | Annotation of text fragments with instances of concepts, rather than concepts themselves | Supports free text annotations classifiable by a predefined set of categories. |
| MemoNote | Three different types of ontologies: pedagogy annotation objectives ontology, a domain annotation objectives ontology and document annotation objectives ontology | Provides some support to semi-automatic annotation by means of annotation patterns |
| AeroDAML | DAML ontologies | Supports annotation with customized ontologies |
| AKTiveMedia | Ontology-based and free-text annotations | Annotating documents with support of text, images and HTML documents (containing both text and images) |

| Tool | Ontologies | Other features |
|------|-----------|----------------|
| KIM | Supports both semantic and free text annotations | Content retrieval based on semantic restrictions, and querying and modifying the underlying ontologies and knowledge bases |
| @note | Combines free-text and ontology-based semantic annotations | Facilitates the collaborative definition of critical annotation ontologies among various instructors |

# 6 Discussion

Figure 1 shows a summary of the tools analysed, which shows that 15.15% do not use any annotation classification system, 39.39% use some type of controlled vocabulary, whether it is constituted of semantic tags or style tags, 21.21% use extensible vocabularies e built by the annotators themselves, and finally 24.24% use ontologies to classify the annotations. Therefore, in absolute terms, 84.84% use some system to organise annotations in order to classify them to be able to exploit this information.

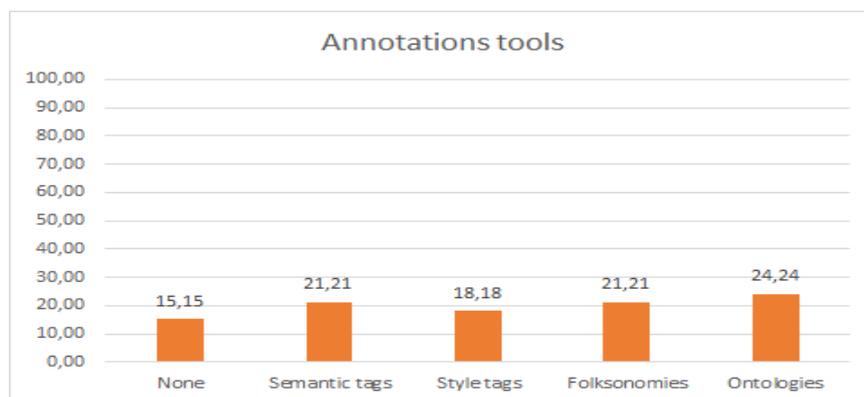

**Fig. 1:** Distribution of analysed annotation tools

From the educational point of view (Kalboussi A et al, 2016), the tools that do not allow the classification of annotations are reduced to a simple simulation of an activity that was previously done by hand. Consequently, the possibility of exploiting the information that can be obtained from the annotation process is lost. In this sense, aspects such as the annotation styles or the annotation models, which could emerge from the analysis of the annotations, are lost, so that neither students nor teachers can benefit from this information.

Concerning the tools that use controlled and uncontrolled vocabularies, it is possible to exploit the information about the annotation process and obtain information about annotation styles. However, the flat structure of these vocabularies is limiting because it is not possible to obtain complex annotation models, since the tags used are unrelated (they are plain sets of keywords that are assigned to the annotations.) The most advanced type of possible exploitation would be to make clusters of terms that are used in certain contents, but it would be impossible to obtain more complex information. From the educational point of view, the use of controlled vocabularies makes it possible to analyse the adequacy for students of the limited set of terms that is provided to categorize a piece of content, showing their ability to combine the terms and describe the contents. However, with uncontrolled vocabularies, another aspect can be measured, which is the creative and reflective capacity to find the most adequate terms to describe a content. The main problem of uncontrolled vocabularies is the open nature of these vocabularies, so that it can be difficult to look for annotation models if very different terms are used, even if they are synonymous.

Finally, there are the tools that use structured vocabularies, and in particular ontologies. From the educational point of view, this is the most flexible option and the one that can provide teachers with most information on the way in which students annotate. Firstly, the use of ontologies makes it possible to use the most appropriate vocabulary for a specific piece of content. And secondly, due to the very structure of an ontology, it is the most effective mechanism for associating information and classifying annotations since the relationships between triplets make it possible to associate not only direct properties but properties that can be deduced from the structure of the ontology. In this sense, the use of ontologies to annotate makes it possible to show a student's reflective capacity and maturity level since he/she will have to think about how to use the complex semantic structure offered by an ontology. In addition to the use of an ontology, the teacher, through many of the tools provided, is able not only to retrieve the annotations, but he/she can profit of other systems that exploit this information such as recommenders, annotation models and even extraction of new ontologies based on the actual use of the conceptual structures (e.g., considering that certain concepts are not used or that the ontology lacks of other relevant concepts).

Annotation tools currently appear within other ecosystems, especially in digital book editing tools or ebooks, so they have become a component of ebooks. In this sense, annotations tend to be "content-rich" Thus, not only do they contain text but they can contain other types of content such as videos (Tanioka et al, 2017), sound (Rich et al, 2008) and images, becoming one more element of the ebook. It is important to highlight the important role they play within what is called interactive fiction. In this area, about the focus is placed on enhancing participation and interaction with the reader in a digital book. Thus annotations are an element widely used in this area to provide information to the reader of the book as part of their interaction (e.g., an interactive fiction entry could contain a map that illustrates information in a book that talks about a city). That is why for this type of content-rich annotations it may be necessary to use more semantic classification systems to exploit the forms of annotation used.

# 7 Conclusions and future work.

This paper provides a review of the content annotation tools from the point of view of the way in which these tools classify annotations. The results show that most tools offer some classification system with the aim of being able to subsequently exploit the information. A classification of existing tools has been proposed according to the structure of the classification system. In this sense, the use of controlled vocabularies, uncontrolled vocabularies and structured vocabularies has been identified. In the latter case, it has been found that, of the possibilities available (taxonomies, thesauri, and ontologies), ontologies are preferably used due to the flexibility they offer, the possibility to better adapting to the type of content to be annotated, and their semantic richness.

As lines for future work, we propose extending this study by also integrating the new content-rich annotation types, as well as considering annotations that are not made on documents.

# Bibliography

Tront, J.G, Eligeti., V., Prey, J. 2006. Classroom Presentations Using Tablet PCs and WriteOn. In: Proceedings of the 36th ASEE/IEEE Frontiers in Education Conference, San Diego, CA, USA, pp. 1-5. doi: 10.1109/FIE.2006.322336

Liao C., Guimbretière F., Anderson R., Linnell N., Prince C., Razmov V. 2007. PaperCP: Exploring the Integration of Physical and Digital Affordances for Active Learning. In: Proceedings of the 11th IFIP Conference on Human-Computer Interaction (INTERACT'07) Part II, Rio de Janeiro, Brazil, pp. 15-28. doi: 10.1007/978-3-540-74800-7_2

Pearson, J., Buchana, G., Thimbleby, H., Jones, M. 2012. The Digital Reading Desk: A lightweight approach to digital note-taking. Interacting with Computers, 24(5), pp. 327-338. doi: 10.1016/j.intcom.2012.03.001

Kam, M., Wang, J., Iles, A., Tse, E., Chiu, J., Glaser, D. 2005. Livenotes: a system for cooperative and augmented note-taking in lectures. In: Proceedings of the SIGCHI Conference on Human Factors in Computing Systems (CHI '05), Portland, Oregon, USA, pp. 531-540. doi: 10.1145/1054972.1055046

Chatti, M.A., Sodhi, T., Specht, M., Klamma, R., Klemke, R. 2006. u-Annotate: An Application for User-Driven Freeform Digital Ink Annotation of E-Learning Content. In: proceedings of the Sixth International Conference on Advanced Learning Technologies (ICALT'06), Kerkrade, The Netherlands, pp. 1039-1043. doi: 10.1109/ICALT.2006.1652624

Pereira, N-B., Kawase R., Dietze S., Bernardino-de-Campos, G.H., Nejdl W. 2012 Annotation Tool for Enhancing E-Learning Courses. In: Proceedings of the 11th International Conference on Web Based Learning (ICWL'12), Sinaia, Romania, pp. 51-60. doi: 10.1007/978-3-642-33642-3_6

Su, A. Y., Yang, S. J., Hwang, W. Y., Zhang, J. 2010. A Web 2.0-based collaborative annotation system for enhancing knowledge sharing in collaborative learning environments. Computers & Education, 55(2), 752-766. doi: 10.1016/j.compedu.2010.03.008

Chen, Y-C, Hwang, R-H., Wang, C-Y. 2012 Development and evaluation of a Web 2.0 annotation system as a learning tool in an e-learning environment. Computers & Education, 58(4), 1094-1105. doi: 10.1016/j.compedu.2011.12.017.

Jan, J. C., Chen, C. M., Huang, P. H., 2016. Enhancement of digital reading performance by using a novel web-based collaborative reading annotation system with two quality annotation filtering mechanisms. International Journal of Human-Computer Studies. 86, 81-93. doi: 10.1016/j.ijhcs.2015.09.006

Bottoni, P., Civica, R., Levialdi, S., Orso, L., Panizzi, E., Trinchese, R. 2004. MADCOW: a multimedia digital annotation system. In: Proceedings of the working conference on Advanced visual interfaces (AVI '04), Gallipoli, Italy, pp. 55-62. doi: 10.1145/989863.989870

Chen, C. M., Chen, F. Y. 2014. Enhancing digital reading performance with a collaborative reading annotation system. Computers & Education, 77, 67-81. doi:10.1016/j.compedu.2014.04.010

Shindo, H., Munesada, Y., & Matsumoto, Y. 2018. PDFAnno: a Web-based Linguistic Annotation Tool for PDF Documents. In Proceedings of the Eleventh International Conference on Language Resources and Evaluation (LREC-2018).

Deng, L., Li, S. C., & Lu, J. 2018. Supporting collaborative group projects with Web 2.0 tools: A holistic approach. Innovations in Education and Teaching International, 55(6), 724-734.

Lu, J. & Deng, L. 2013. Examining students' use of online annotation tools in support of argumentative reading. Australasian Journal of Educational Technology, 29(2), 161-171, doi: 10.14742/ajet.159

Glover, I., Hardaker, G., Xu, Z. 2004. Collaborative annotation system environment (CASE) for online learning, Campus-Wide Information Systems, 21(2), 72-80. doi: 10.1108/10650740410529501

Lin, S.J., Chen, H.Y., Chian, Y-T., Luo, G.H., Yuan, S.M. 2014. Supporting Online Reading of Science Expository with iRuns annotation strategy. In: Proceedings of the 7th International Conference on Ubi-Media Computing (U-MEDIA'14), Ulaanbaatar, Mongolia, pp 309-321. doi: 10.1109/U-MEDIA.2014.59

Hwang,W-Y., Wang, C-H., Sharples, M. 2007. A study of multimedia annotation of Web-based materials. Computers & Education, 48, 680–699. doi: 10.1016/j.compedu.2005.04.020

Asai, H., Yamana, H. 2014. Intelligent Ink Annotation Framework that uses User's Intention in Electronic Document Annotation. In: Proceedings of the Ninth ACM International Conference on Interactive Tabletops and Surfaces (ITS '14), Dresden, Germany, pp. 333-338. doi: 10.1145/2669485.2669542

Chang, M.H., Kuo, R., Chang, M., Kinshuk, Kung, H-Y. 2015. Online Annotation System and Student Clustering Platform. In: Proceedings of the 8th International Conference on Ubi-Media Computing (UMEDIA'15), Colombo, Sri Lanka, pp. 202-207. doi: 10.1109/UMEDIA.2015.7297455

Lebow, D.G & Lick, D.W. HyLighter: An Effective Interactive Annotation Innovation for Distance Education. 2005. In: Proceedings of the 21th Annual Conference on Distance Teaching and Learning, Madison Wisconsin, 5pp.

Anagnostopoulou, C. & Howell, F. 2009. Collaborative online annotation of musical scores for eLearning using A.nnotate.com. In: Proceedings of 3rd International Technology, Education and Development Conference (INTED'09), Valencia, Spain, pp. 4071-4079

Paradis, J. & Fendt, K. 2016. Annotation Studio - Digital Annotation as an Educational Approach in the Humanities and Arts. Tech. Report 110196. MIT

Tseng, S-S., Yeh, H-C., Yang, S-H. 2015 Promoting different reading comprehension levels through online annotations. Computer Assisted Language Learning, 28(1), 41-57. doi: 10.1080/09588221.2014.927366

Kim, J-K., Sohn, W-S., Hur, K. 2014. Increasing learning effect by tag cloud interface with annotation similarity. International Journal of Advanced Media and Communication, 5(2/3), 135-148. doi: 10.1504/IJAMC.2014.060503

Bateman, S., Brooks, C., McCalla, G., Brusilovsky, P. 2007. Applying Collaborative Tagging to E-Learning. In: Proceedings of the 6th International World Wide Web Conference (WWW'07), Banff, Alberta, Canada, 7pp.

Baterman, S., Farzan, R., Brusilovsky, P., McCalla., G. 2006. OATS: The Open Annotation and Tagging System. In: Proceedings of the 3rd Annual Scientific Conference of the LORNET Research Network (I2LOR'2006), Montreal, Canada. 10pp.

Kawase, R., Herder, E., Nejdl, W. 2009. A comparison of paper-based and online annotations in the workplace. In: Proceedings of the European Conference on Technology Enhanced Learning (EC-TEL), Niza, France, pp. 240-253. doi: 10.1007/978-3-642-04636-0_23

Kawase, R., Herder, E., Nejdl, W. 2010. Annotations and Hypertrails With SpreadCrumbs - An Easy Way to Annotate, Refind and Share. In Proceedings of 6th International Conference on Web Information Systems and Technologies (WEBIST'10) Valencia, Spain, 8pp.

Silvestre, F., Vidal, P., Broisin, J. 2014. Tsaap-Notes – An Open Micro-Blogging Tool for Collaborative Notetaking during Face-to-Face Lectures. In: Proceedings of the 2014 IEEE 14th International Conference on Advanced Learning Technologies (ICALT'14), Athens, Greece, pp 39-43. doi: 10.1109/ICALT.2014.22

Hinze, A., Heese, R., Schlegel, A., Luczak-Rösch, M. 2012. User-Defined Semantic Enrichment of Full-Text Documents: Experiences and Lessons Learned. In: Proceedings of the 16th International Conference on Theory and Practice of Digital Libraries (TPDL'12), Paphos, Cyprus, pp. 209-214. doi: 10.1007/978-3-642-33290-6_23

Luczak-Rösch, M, & Heese, R. 2009. Linked Data Authoring for Non-Experts. In: Proceedings of the 18th International World-Wide-Web Conference (WWW2009), Madrid, Spain. 5pp.

Azouaou, F & Mokeddem, H. 2013. WebAnnot: a learner's dedicated web-based annotation tool. International Journal of Technology Enhanced Learning, 5(1), 56-84. doi: 10.1504/IJTEL.2013.055949

Rocha, T-R., Willrich, R., Fileto, R., Tazi, S. 2009. Supporting Collaborative Learning Activities with a Digital Library and Annotations. In: Proceedings of the 9th IFIP TC 3 World Conference on Computers in Education (WCCE 2009), Bento Gonçalves, Brazil, pp. 349-358. doi: 10.1007/978-3-642-03115-1_37

Azouaou, F. & Desmoulins, C. 2006. Teachers' document annotating: models for a digital memory tool. International Journal of Continuing Engineering Education and Life Long Learning, 16(1-2), 18-34. doi: 10.1504/IJCEELL.2006.008915

Azouaou, F. & Desmoulins; C. 2006. MemoNote, a context-aware annotation tool for teachers. In: Proceedings of the 7th International Conference on Information Technology Based Higher Education and Training (ITHET '06), Sydney, NSW.pp. 621-628. doi: 10.1109/ITHET.2006.339677

P. Kogut and W. Holmes, "AeroDAML: Applying Information Extraction to Generate DAML Annotations from Web Pages," in First International Conference on Knowledge Capture KCAP 2001 Workshop on Knowledge Markup and Semantic Annotation, 2001, vol. 21, p. 3.

A. Chakravarthy, F. Ciravegna, and V. Lanfranchi, "Cross-media document annotation and enrichment," in Proc. 1st Semantic Web Authoring and Annotation Workshop (SAAW2006), 2006.

N. Bikakis, G. Giannopoulos, T. Dalamagas, and T. Sellis, "Integrating keywords and semantics on document annotation and search," On the Move to Meaningful Internet Systems, OTM 2010, pp. 921–938, 2010.

B. Popov, A. Kiryakov, D. Ognyanoff, D. Manov, and A. Kirilov, "KIM–a semantic platform for information extraction and retrieval," Natural Language Engineering, vol. 10, no. 3–4, pp. 375–392, 2004.

S. K. Malik, N. Prakash, and S. Rizvi, "Semantic Annotation Framework For Intelligent Information Retrieval Using KIM Architecture," International Journal, vol. 1, no. October, pp. 12–26, 2010.

B. Popov, A. Kiryakov, A. Kirilov, and D. Manov, "KIM – Semantic Annotation Platform," Engineering, pp. 834–849, 2003.

S. Handschuh, S. Staab, and F. Ciravegna, "S-CREAM—semiautomatic creation of metadata," Knowledge Engineering and Knowledge Management: Ontologies and the Semantic Web, pp. 165–184, 2002.

S. Corlosquet, R. Delbru, and T. Clark, "Produce and Consume Linked Data with Drupal!," The Semantic Web- …, vol. 1380, pp. 751–766, 2009.

R. Shreves, "Open source cms market share. White paper, Water & Stone," 2011.

Chowdhury, G., & Chowdhury, S. 2002. Introduction to digital libraries. Facet publishing

Schroeter, R., Hunter, J., Guerin, J., Khan, I., & Henderson, M. 2006. A synchronous multimedia annotation system for secure collaboratories. In: Proceedings of the Second IEEE International Conference on e-Science and Grid Computing

Kawase, R., Herder, E., Nejdl, W. 2009. A comparison of paper-based and online annotations in the workplace. In: Proceedings of the European Conference on Technology Enhanced Learning (EC-TEL), Niza, France, pp. 240-253. doi: 10.1007/978-3-642-04636-0_23

Jan, J. C., Chen, C. M., Huang, P. H., 2016. Enhancement of digital reading performance by using a novel web-based collaborative reading annotation system with two quality annotation filtering mechanisms. International Journal of Human-Computer Studies. 86, 81-93. doi: 10.1016/j.ijhcs.2015.09.006

Kawasaki, Y., Sasaki, H., Yamaguchi, H., Yamaguchi, Y., 2008. Effectiveness of highlighting as prompt in text reading on computer monitor. In: Proceedings of the 8th WSEAS International Conference on Multimedia Systems and Signal Processing (MUSP), Hangzhou, China, pp. 100-112.

Chen, C. M., Chen, F. Y. 2014. Enhancing digital reading performance with a collaborative reading annotation system. Computers & Education, 77, 67-81. doi:10.1016/j.compedu.2014.04.010

Hwang,W-Y., Wang, C-H., Sharples, M. 2007. A study of multimedia annotation of Web-based materials. Computers & Education, 48, 680–699. doi: 10.1016/j.compedu.2005.04.020

Johnson, T. E., Archibald, T. N., Tenenbaum, G. 2010. Individual and team annotation effects on students' reading comprehension, critical thinking, and meta-cognitive skills. Computers in Human Behavior, 26(6), 1496–1507. doi: 10.1016/j.chb.2010.05.014

Mendenhall, A., & Johnson, T. E. 2010. Fostering the development of critical thinking skills, and reading comprehension of undergraduates using a Web 2.0 tool coupled with a learning system. Interactive Learning Environments, 18(3), 263–276. doi: 10.1080/10494820.2010.500537

Nokelainen, P., Kurhila, J., Miettinen, M., Floréen, P., Tirri, H. 2003. Evaluating the role of a shared document-based annotation tool in learner-centered collaborative learning. In: Proceedings of the 3rd IEEE International Conference on Advanced Learning Technologies (ICALT'03), Athens, Greece, pp. 200-203. doi: 10.1109/ICALT.2003.1215056

Nokelainen, P., Miettinen, M., Kurhila, J., Floréen, P., Tirri, H. 2005. A shared document-based annotation tool to support learner-centred collaborative learning. British Journal of Educational Technology, 36(5), 757–770. doi:10.1111/j.1467-8535.2005.00474.x.

Su, A. Y., Yang, S. J., Hwang, W. Y., Zhang, J. 2010. A Web 2.0-based collaborative annotation system for enhancing knowledge sharing in collaborative learning environments. Computers & Education, 55(2), 752-766. doi: 10.1016/j.compedu.2010.03.008

Cigarrán-Recuero, J., Gayoso-Cabada, J., Rodríguez-Artacho, M-A., Romero-López, D., Sarasa, A., Sierra, J-L. 2014. Assessing semantic annotation activities with formal concept analysis. Expert Systems with Applications, 41(11), 5495-5508. doi: 10.1016/j.eswa.2014.02.036

Buendía, F. 2016. Design of an annotation tool for educational resources. In: Proceedings of the Fourth International Conference on Technological Ecosystems for Enhancing Multiculturality (TEEM '16), 1005-1009. doi: 10.1145/3012430.3012639

Kalboussi A., Omheni N., Mazhoud O., Kacem A.H. 2015. How to Organize the Annotation Systems in Human-Computer Environment: Study, Classification and Observations. In: Proceedings of the 15th IFIP TC 13 International Conference on Human-Computer Interaction (INTERACT'15). doi: https://doi.org/10.1007/978-3-319-22668-2_11

Kalboussi, A., Mazhoud, O., Kacem, A-H. 2016. Comparative study of web annotation systems used by learners to enhance educational practices: features and services. Int. J. Technology Enhanced Learning, 8(2), 129-150. doi: 10.1504/IJTEL.2016.078081

Novak, E., Razzouk,R, Johnson, J-E. 2012. The educational use of social annotation tools in higher education: A literature review. The Internet and Higher Education 15(1), 39-49. doi: 10.1016/j.iheduc.2011.09.002.

Gayoso-Cabada, J., Sarasa-Cabezuelo, A., & Sierra, J. L. 2018. Document Annotation Tools: Annotation Classification Mechanisms. In Proceedings of the Sixth International Conference on Technological Ecosystems for Enhancing Multiculturality (pp. 889-895). ACM.

Ovsiannikov, I.A., Arbib, M.A., McNeill, T-H. 1999. Annotation technology. International Journal of Human-Computer Studies 50(4), 329-362. 10.1006/ijhc.1999.0247

Tanioka, T., Okuno, K., Gokita, K., Nakagawa, M., Inokuchi, M., Kojima, K. 2017. An Intraoperative Educational Annotation System For Endoscopic Surgery, Journal of the American College of Surgeons, 225(4), e42. doi: 10.1016/j.jamcollsurg.2017.07.630.

P-J. Rich & M. Hannafin. 2008. Video Annotation Tools. Technologies to Scaffold, Structure, and Transform Teacher Reflection. Journal of Teacher Education, 60(1), 52-67. doi: 10.1177/0022487108328486.